\documentclass[letterpaper, 10 pt, conference]{ieeeconf}  
\usepackage{cite}    
\usepackage{color}
\usepackage{balance}
\usepackage{graphicx}
\usepackage[bookmarks=false]{hyperref}

\IEEEoverridecommandlockouts                              
\newcommand{\ie}{\textit{i.e.,}}

\title{\LARGE \bf
TFCheck : A TensorFlow Library for Detecting Training Issues in Neural Network Programs}

\author{Houssem Ben Braiek and Foutse Khomh \\
\emph{SWAT Lab., Polytechnique Montr\'{e}al, Montr\'{e}al, Canada}
    \\
     \emph{\{houssem.ben-braiek, foutse.khomh\}@polymtl.ca}
    }

\begin{document}

\maketitle

\thispagestyle{empty}
\pagestyle{empty}


\begin{abstract}

The increasing inclusion of Machine Learning (ML) models in safety critical systems like autonomous cars have led to the development of multiple model-based ML testing techniques. One common denominator of these testing techniques is their assumption that training programs are adequate and bug-free. These techniques only focus on assessing the performance of the constructed model using manually labeled data or automatically generated data. However, their assumptions about the training program are not always true as training programs can contain inconsistencies and bugs. In this paper, we examine training issues in ML programs and propose a catalog of verification routines that can be used to detect the identified issues, automatically. We implemented the routines in a  Tensorflow-based library named TFCheck. Using TFCheck, practitioners can detect the aforementioned issues automatically. To assess the effectiveness of TFCheck, we conducted a case study with real-world, mutants, and synthetic training programs. Results show that TFCheck can successfully detect training issues in ML code implementations.

\end{abstract}

\section{INTRODUCTION}
Nowadays, software applications powered by Machine Learning (ML) are increasingly being deployed in safety-critical systems such as self-driving cars or aircraft collision-avoidance systems. Therefore, their reliability is now of paramount importance. Recently, researchers have proposed many testing approaches to help improve the reliability of ML applications~\cite{OurPaper}. A common denominator of these testing approaches is the evaluation of the model performance in terms of prediction ability regarding manually labeled data and--or automatically generated data set. The test data set is used to check for inconsistencies in the behavior of the models. Whenever inconsistencies are uncovered, the training set is augmented with miss-classified test data in order to help the model learn the properties of corner-cases on which it performed poorly. This process is repeated until a satisfactory performance is achieved. These proposed ML testing approaches assume that the ML model is trained adequately, i.e., the training program is bug-free and numerically stable. They also assume that the training algorithm and the model hyper-parameters are optimal in the sense that the model has the adequate capacity to learn the patterns needed to perform the targeted task, adequately. However, bugs exist in ML training codes and these bugs can invalidate some of these assumptions. In fact, Zhang et al.~\cite{TFBugs} investigated bugs in neural networks training programs built on TensorFlow\cite{cai2016tensorflow} and reported about multiple occurrences. They also identified five challenges related to bug detection and localization. One of these challenges is coincidental correctness. A coincidental correctness occurs when a bug exists in a program, but by coincidence, no failure is detected. Coincidental correctness can be caused by undefined values such NaNs and Infs, induced by numerically unstable functions. Finding training input data to expose these issues can be challenging. Also, a bug in the implementation of a neural network can result in saturated or untrained neurons that do not contribute to the optimization, preventing the model from learning properly. Moreover, when a neural network makes mistakes on some adversarial data, gathering more data is not a panacea. The neural network model may not have the appropriate capacity to learn patterns from these noisy data or may miss good regularization to avoid overfiting the noise. Detecting all these issues requires effective verification mechanisms. In this paper, we introduce a list of verification routines to help developers detect and correct errors in their ML training programs. Since these verification routines can be difficult to adopt for beginners and since their manual application can be very time-consuming, we developed TFCheck, a TensorFlow (TF) library that implements the proposed verification routines. To assess the effectiveness of TFCheck at detecting bugs in ML training programs automatically, we conducted a case study using real-world, mutants and synthetic training programs. Results of the case study show that using TFCheck, developers can successfully detect training issues in a ML code implementation. This paper makes the following contributions:
\begin{itemize}
\item We provide a practical guide outlining issues and verification routines, that developers can use to detect and correct issues in their ML training programs. 
\item We also provide TFCheck, a TF-based library implementing the verification routines presented in our guide. 
Using TFCheck, developers can monitor and debug TF programs automatically. 
\end{itemize}

\textbf{The remainder of this paper is organised as follows.} Section~\ref{background} provides background information about Deep Neural Networks (DNNs) and TF. Section~\ref{relatedWork} discusses the related literature. Section~\ref{heuristics} presents some common issues experienced by developers when training ML models alongside some verification mechanisms. Section~\ref{library} describes the structure of our TF-based library TFCheck and its utilization. Section~\ref{evaluation} reports about a case study aimed at evaluating TFCheck. Finally, Section~\ref{conclusion} concludes the paper. 

\section{Background}\label{background}
This section introduces key concepts of the paper. 
\subsection{Deep Neural Network}
Neural Networks (NN) are powerful statistical learning models that can solve complex classification and regression problems. 
A NN is composed of 
interconnected layers of computation units, mimicking the structures of brain's neurons and their connectivity. 
Each neuron includes an activation function (\ie{} a non-linear transformation) to a weighted sum of input values with bias. The predicted output of a NN is calculated by computing the outputs of each neuron through the network layers in a feed-forward manner. 
In fact, DNNs, which are the backbone of deep learning use a cascade of multiple hidden layers to increase exponentially, the learning capacity of NNs. 
DNNs are built using 
differentiable model fitting, which is an iterative process during which the model trains itself on input data through gradient-based optimization routines, making small 
adjustments iteratively, with the aim of refining the model until it predicts mostly right outputs.
The principal components of a NN are: \\
\textbf{Parameters} represent the trained weights and biases used by neurons to make their internal calculations.\\ 
\textbf{Activations} represent non-linear functions that add non-linearity to neuron output aimed at indicating, fundamentally, if a neuron should be activated or not.\\
\textbf{Loss Function} consists of a math function, which estimates the distance between predicted and actual outcomes. 
If the DNN's predictions are perfect, the loss is zero; otherwise, the loss is greater than zero. \\
\textbf{Regularization} consists in techniques that penalize the model's complexity to prevent overfitting such as $L1$-$L2$ regularization or \emph{dropout}.\\
\textbf{Optimizer} adjusts the parameters of the model iteratively (reducing the objective function), 
in order to: (1) build the best-fitted model \ie{} lowest loss; and (2) keep the model as simple as possible \ie{} strong regularization. The most used optimizers are based on gradient descent algorithms.\\
\textbf{Hyperparameters} are model's parameters that are constant during the training phase and which can be fixed before running the fitting process such as the number of layers or the learning rate. 

When training a DNN, ML developers set hyperparameters, choose loss functions, regularization techniques, and gradient-based optimizers, following best practices or guidelines derived from the literature. After training the model, the best-fitted model is evaluated on a testing dataset (which should be different from the training dataset), using error rate and accuracy measures. The occurrence of errors in the training program of a DNN often translates into poor model performance. Therefore, it is important to ensure that DNN program implementations are bug-free. Given the large size of the testing space of a DNN~\cite{OurPaper}, systematic debugging and testing techniques are required to assist developers in errors detection and correction activities. 

\subsection{TensorFlow Framework}
TensorFlow (TF) is a popular library that can be leverage to create DNNs. 
TF is based on a 
Directed Acyclic Graph (DAG) that contains nodes, which represent mathematical operations, and edges, which encapsulate tensors (i.e., multidimensional data arrays). This computational graph offers a high-level of abstraction to represent the algebraic computation of the constructed DNN models, independently of the programming language used, the execution environment, and the target hardware. Thus, it provides TF users with the flexibility to execute their computation on one or more CPUs/GPUs, or even mobile devices. 

TF uses a concept known as deferred execution or lazy evaluation. This means that there are two principal phases in a TF program: (1) a construction phase, that assembles a graph, including variables and operations ; (2) an execution phase that uses a session to execute operations and evaluate results in the graph. 
The constructed DAG allows the TF XLA compiler to generate an optimized and faster code for any targeted deployment environment. Indeed, the TF execution engine schedules the designed computations' tasks in a way that ensures the efficient use of resources. 
In our work, we choose to implement our verification routines on TF because of its high popularity in the ML community\cite{braiek2018open}. The communication interface between a TF model under test and our TFCheck library is its corresponding DAG.

\section{Related work}\label{relatedWork}
Roger et al.~\cite{MCMCTesting} applied property-based testing to detect errors in implementations of MCMC (Markov chain Monte Carlo) samplers. Property-based testing is a technique that consists in inferring the properties of a computation using the theory and formulating invariants that should be satisfied by the program. Using the formulated invariants, test cases are generated and executed repeatedly throughout the computation to detect inconsistencies in the system. Thus, to test the correctness of MCMC using property-based approach, Roger et al. verified that the conditional distribution was consistent with the joint distribution for the produced samples at each update iteration. In order words, instead of comparing the outputs with a reference oracle (that could be not available in such case), one can ensure that probability laws hold throughout the execution of a scientific program. For our goal of testing the sanity of DNN training program, we assume that almost all the mathematical functions used are provided as ready-to-use routines by ML libraries. The target issues are related to the training program that assembles those components as the DNN implementation and sets the different hyperparameters as the DNN configuration. Therefore, most of our verification routines are rooted in 
fundamental statistical and mathematical concepts that could indicate the presence of inappropriate configuration or implementation errors. Selsam et al.~\cite{bugfreeMLS} proposed to formally specify the computations of ML programs and constructed formal proofs of written theorems that define what it means for a program to be correct and error-free. Using these formal mathematical specifications 
and a machine-checkable proof of correctness representing a formal certificate that can be verified by a stand-alone machine without human intervention, they analyzed a ML program designed for optimizing over stochastic computation graphs, using the interactive proof assistant \textit{Lean}, and reported it to be bug-free. In this paper, we propose a series of verification operations that  can be used to detect errors in the implementation of ML training programs. 

The TF official 
debugging tool (tfdbg)~\cite{tensorflowDebugger} offers features such as DAG inspection, real-time tensor values, an conditional breakpoints.
However, it is not very practical since it adds a huge overhead on computation time, as it handles each execution step of the graph, to allow TF user to debug the issues and pinpoint the exact graph nodes where a problem first surfaced. 
Thus, TF users often rely on Tensorboard visualization tool to interactively display the curves of measures and the data flow structure of the running DAG~\cite{wongsuphasawat2018visualizing}. 
They use named scope to define hierarchy on the graph to have nodes' names like “dense\_layer/weight” pr “dense\_layer/Relu”. This is useful to collapse the variables from one scope during the visualization of multiple layers. We intend to build our TFCheck library following this established naming convention and the simple use of session object that provides a connection between the Python code and the DAG on execution. Through this object, it is possible to fetch any graph node value by its predefined name. Hence, 
TFCheck performs verification operations 
on the values of nodes fetched before and after running training operations. These verification operations 
do not require to break neither the feed-forward nor the backward pass. Therefore, TFCheck introduces the verification routines before and after training operation runs while keeping the training pass to be executed in an atomic way.
\section{ML model training pitfalls}\label{heuristics}
Many issues can prevent a proper training of a DNN. Below, we elaborate on some of these issues divided into three groups based on the affected components: Parameter-related, Activation-related, and Optimization-related issues.
\subsection{Parameters related issues}
In the following, we discuss three different types of issues that are related to parameters and their verification routines. 
\subsubsection{Untrained Parameters}
The most basic and common issue in relation to model's parameters occurs when developers forget to connect the different branches of a DNN or call the function of the creation layer with incorrect parameters. A DNN can still train and converge with less number of layers, but it could not have enough learning capacity to reach high prediction ability. Therefore, it is important to check that all defined parameters are modified appropriately. 
\textbf{Verification routine.} This issue can be detected by storing each actual tensor's parameter and comparing it with the tensor's parameters obtained after the execution of training operations, in order to make sure that each parameter is getting optimized and differs from its default value. 

\subsubsection{Poor Weight Initialization} 
The initialization of weights values should be done carefully. 
Weights need to be initialized in a way that breaks the symmetry between hidden neurons of the same layer, because if hidden units of the same layer share the same input and output weights, they will compute the same output and receive the same gradient, hence performing the same update and remaining identical, thus wasting capacity. In other words, there is no source of asymmetry between neurons if their weights are initialized to be the same. The initial random weights values should be small enough to neither diverge while the gradients explode, nor very close to zero, so the gradients vanish quickly.\\ 
\textbf{Verification routine. } One can verify that there are significant differences between the parameter's values by computing the variance of each parameter's values and checking if it is not equal or very close to $0$.
\subsubsection{Parameters' Values Divergence} Weights can diverge to $+/-\infty$ if the initial values or the learning rate are too high and there is a lack of--or--insufficient regularization, because the back-propagation updates process would push the weights to becoming higher and higher, until reaching $\infty$ values (this is caused by overflow rounding precision). In addition to that, biases also risk divergence in the sense that they can become huge in certain situations where features could not explain enough the predicted outcome and might not be useful in differentiating between available classes. This problem occurs more likely when the distribution of classes is very imbalanced.\\
\textbf{Verification routine. } One can verify that parameters' values are not diverging by computing their $75^{th}$ percentiles (upper quartiles) and verifying that they are less than a predefined threshold, during each training iteration.

\subsubsection{Parameters' Unstable learning} The use of several hidden layers can cause unstable learning situations such as: (1) Layer's parameters changing rapidly in an unstable way; preventing the model from learning relevant features. The intuition is that the parameters are a part of the estimated mapping function, so we risk overfitting the current processed batch of data when we try to adapt strongly the parameters in order to fit this batch; (2) Layer's parameters changing slowly; making it difficult 
to learn useful features from data. These learning issues are strongly related to the unstable gradient phenomenon. Indeed, we mentioned that the computation of the gradients with respect to earlier layers contains a product of terms from all the later layers. This backpropagation of gradient tend to establish significantly different learning speeds in the layers. To ensure stability, advanced mechanisms or adequate hyperparameters choices are needed to establish relatively similar learning speed with respect to all neural network's layers, through balancing out the computed gradients with respect to the layers' parameters as much as possible. In fact, the parameters' updates are computed by an optimizer using not only the gradient and the learning rate. It also contains other hyper-parameters such as momentum coefficients to provide adaptive gradient steps.\\ 
As suggested by Bottou~\cite{leon2015}, it is useful to compare the magnitude of parameter gradients to the magnitude of the parameters themselves with the aim of verifying that the magnitude of parameter updates over batches represent something like $1\%$ of the magnitude of the parameter, not $50\%$ or $0.001\%$.\\
\textbf{Verification routine.} This issue can be detected by comparing the magnitude of parameters' gradients to the magnitude of the parameters themselves. More specifically, following Bottou's recommendation to keep the parameter update ratio around $0.01$ (i.e., $-2$ on base 10 logarithm), one can compute the ratio of absolute average magnitudes of these values and verify that this ratio doesn't diverge significantly from the following predefined thresholds (see the inequation \ref{ineq:update_ratio}).
\begin{equation}
\label{ineq:update_ratio}
-4 < \log_{10}\left(\frac{mean(abs(updates))}{mean(abs(parameters))}\right) < -1
\end{equation}
Using this verification routine, one can recognize the layers where updates seem to be unstable or stalled. 
\subsection{Activation related issues}
In the following, we detail three categories of problems related to activation and their verification routines.
\subsubsection{Activations Out of Range}
Activation functions have specific output range. One common mistake is to implement a mathematical function or to apply a wrong existing function for a layer assuming inaccurate outputs ranges. It is necessary to make sure that activation's range of values stays consistent with what is expected theoretically from their corresponding function (e.g., sigmoid's outputs are between $[0,1]$ and  tanh's outputs are between $[-1,1]$).\\
\textbf{Verification routine.}
To detect activation out of range issues, one can verify if 
the activation values exceed known theoretical range of values. This verification routine is particular useful to find computation mistakes when a new activation function is implemented from scratch.
\subsubsection{Neuron Saturation}
Bounded activation functions with a sigmoidal curve, such as sigmoid or tanh, exhibit smooth linear behaviour for inputs within the active range, and become very close to either the lower or the upper asymptotes for relatively large positive and negative inputs. The phenomenon of neuron saturation occurs when a neuron returns almost only values close to the asymptotic limits of the activation functions. In such case, any change on weights of this neuron will not have a noticeable influence on the output of the activation function. As a result, the training process may stagnate with stable parameters, preventing the training algorithm from refining them.\\
\textbf{Verification routine.}
To detect neuron saturation issues in NNs, one can use the single-valued saturation measure $\rho_B$ proposed by Rakitianskaia and Engelbrecht~\cite{saturationNN}. This measure is computed using the outputs of an activation function and is applicable to all bounded activation functions. It is independent of the activation function output range and allows a direct statistical measuring of the degree of saturation between NNs. $\rho_B$ is bounded and easy to interpret: it tends to $1$ as the degree of saturation increases, and tends to zero otherwise. It contains a single tunable parameter, the number of bins $B$, that converges for $B \geq 10$, i.e., it means splitting the interval of activation outputs into $B$ equal sub-intervals. Thus, $B = 10$ can be used without any further tuning. Given a bounded activation function $g$, $\rho_B$ is computed as the weighted mean presented in Equation~\ref{rhoB}.
\begin{equation}
\label{rhoB}
\rho_B = \frac{\sum^B_{b=1}|\bar{g}'_b|N_B}{\sum^B_{b=1}N_B}
\end{equation}
Where, $B$ is the total number of bins, $\bar{g}'_b$ is the scaled average of output values in the bin $b$ within the range $[-1,1]$, $N_b$ is the number of outputs in bin $b$. Indeed, this weighted mean formula turns to a simple arithmetic mean when all weights are equal. Thus, if $\bar{g}'_b$ is uniformly distributed in $[−1, 1]$, the value of $\rho_B$ will be close to $0.5$, since absolute activation values are considered, thus all $\bar{g}'_b$ values are squashed to the $[0, 1]$ interval. For a normal distribution of $\bar{g}'_b$, the value of $\rho_B$ will be smaller than $0.5$. The higher the asymptotic frequencies of $\bar{g}'_b$, the closer $\rho_B$ will be to $1$.\\
This verification routine can be automated by storing for each neuron its last $O$ outputs values in a buffer of a limited size. Then, computing its $\rho_B$ metric based on those recent outputs. If the neuron corresponding value tends to be $1$, the neuron can be considered as saturated. After checking all neurons for saturation, one can compute 
the ratio of saturated neurons per layer to alert developers about layers with saturation ratios that surpass a predefined threshold. 
\subsubsection{Dead ReLU}
ReLU stands for rectified linear unit, and is currently the most used activation function in deep learning models, especially CNNs. In short, ReLU is linear (identity) for all positive values, and zero for all negative values. Contrary to other bounded activation functions like sigmoid or tanh, ReLU does not suffer from the saturation problem, because the slope does not saturate when $x$ gets large and the problem of vanishing gradient is less observed when using ReLU as activation function. However, the fact that they are null for all negative values increases the risk of ``dead ReLUs''. A ReLU neuron is considered ``dead'' when it always outputs zero. Such neurons do not have any contribution in identifying patterns in the data nor in class discrimination. Hence, those neurons are useless and if there are many of them, one may end up with completely frozen hidden layers doing nothing. This problem often occur when the learning rate is too high or when there is a large negative bias. Recent ReLU variants such as leaky ReLU and ELU are recommended as good alternatives when lower learning rates do not solve the problem.\\
\textbf{Verification routine.}
A given neuron is considered to be dead if it only returns zero or almost zero value as output. Hence, dead ReLUs issues can be detected by storing 
the last obtained outputs for each neuron in a limited size buffer and checking if almost all the stored values are zero or closer to zero. Whenever we find zero or close to zero values, we can mark the neuron as dead. By computing the ratio of dead neurons per layer, one can detect layers with a high number of dead neurons with respect to a predefined threshold in order to warn developers about these frozen layers.
\subsection{Optimization related issues}
In the following, we present six kinds of issues related to parameters and their verification routines.
\subsubsection{Unable to fit a small sample}
Given a small data set, a DNN model should be able to fit it without any issue, because even simple models would be able to fit a tiny subset of data, otherwise, there is likely a poor configuration or a software defect. In such circumstance, there is no need to try training the DNN on large data for multiple days running.\\
\textbf{Verification routine.} One can verify that the optimization mechanism is working well by performing the training process over a controlled sample data set. Following the recommendations of Karpathy\cite{OurPaper}, one can turn off regularization techniques and train the DNN on a tiny subset of data (i.e., a few data points for each class), and then verify that it achieves zero loss under these circumstances. A failure to achieve this performance would signal a minimization problem. 
\subsubsection{Zero loss }
The training process of DNNs relies on data minimization to adapt optimally the neurons parameters, in a way that fits well the training distribution, while being capable of generalization 
on unseen data provided from the same distribution. When the loss minimization process reaches a zero value, it is an indication that the model overfits the training data, and that its prediction ability could be low on unseen data. 
The optimization process aims to learn the high-level patterns in order to ensure a good differentiation between the classes. \\ 
\textbf{Verification routine.} To detect zero loss issues, Goodfellow et al.~\cite{Goodfellow-et-al-2016} recommend to verify the training program directly, as it is (including regularization), checking that it sufficiently fits the controlled data sample, while keeping its loss different from zero. 
\subsubsection{Slow or Non decreasing loss}
After multiple training iterations, if the model still has non-decreasing or a very slow decreasing loss, it is likely to display a poor performance because of a low learning capacity. This low learning capacity issue could be due to a  
deficiency of the optimizer or an implementation mistake. \\
\textbf{Verification routine.} One can verify that the loss is smoothly decreasing by computing the decreasing loss rate (\ref{eq:loss_rate}) and verifying that this rate is mostly higher than a predefined threshold.
\begin{equation}
\label{eq:loss_rate}
loss\_rate = \frac{current\_loss\_value}{previous\_loss\_value}
\end{equation}
\subsubsection{Diverging loss}
Due to a high learning rate or inadequate loss function, the minimization process can transformed to a maximization process, in the sense that the loss value can start increasing wildly at certain point.\\
\textbf{Verification routine.} To detect diverging loss issues, one should keep track of the loss to make sure that it doesn't start increasing after a certain number of iterations. This verification can be done automatically by initializing and updating a reference variable containing $lowest\_loss\_value$. Using this variable, one can compute the absolute loss rate (\ref{eq:abs_loss_rate}) to check if the last computed losses are diverging from this previously obtained minimum value.
\begin{equation}
\label{eq:abs_loss_rate}
abs\_loss\_rate = \frac{current\_loss\_value}{lowest\_loss\_value}
\end{equation}
\subsubsection{Loss fluctuations}
A highly fluctuating loss during the training phase of a DNN may indicate an inappropriate learning rate that prevents the convergence of gradient-based optimizers to the minimum and keeps jumping it with relatively large updates. Besides, mini-batch stochastic gradient-based optimizers risk encountering a fluctuating loss when the batch size is relatively small, so the loss would be noisy without an evolution trend during multiple iterations.\\
\textbf{Verification routine.} Fluctuations in the loss functions can be identified by the presence of successive loss increases and decreases; so the loss rate (\ref{eq:loss_rate}) alternates between values that are either higher and lower than $1$.
\subsubsection{Unstable Gradients}
Due to poor weights initialization and bad choices of hyperparameters, DNNs can be exposed to the unstable gradient problem, which could manifested in the form of vanishing or exploding gradients as described below.\\
\textit{\textbf{Vanishing Gradient:} }
In this case, the gradient tends to have smaller values when it is back-propagated through the hidden layers of the DNN. This causes the gradient to be almost zero in the earlier layers and consequently would be transformed to undefined values such as Not-a-Number(NaN) caused by underflow rounding precision during discrete executions on hardware. The problem of vanishing gradient can lead to the stagnation of the training process and eventually causing a numerical instability. As an illustration, we take the example of a DNN configured to have sigmoid as activation function and a randomly initialized weight using a Gaussian distribution with a zero mean and a unit standard deviation. The sigmoid function returns a maximum derivative value of $0.25$ (i.e., the derivative of sigmoid when the input is equal to zero) and the absolute value of the weights product is less than $0.25$ since they belong to a limited range between $[-1, 1]$. Generally, the gradient of a given hidden layer in a NN would be computed by the sum of products of all the gradients belonging to the deeper layers and the weights assigned to each of the links between them. As a result, the deeper the gradient is back-propagated over the hidden layers, the more there are products of several terms that are less or equal to $0.25$. Hence, it is apparent that earlier hidden layers (i.e., closer to the input layer) would have very less gradient and would be almost stagnant with less weights changes during the training process.\\
\textit{\textbf{Exploding Gradient:}}
The exploding gradient phenomenon can be encountered when, inversely, the gradient with respect to the earlier layers diverges and its values become huge. As a consequence, this could result in the appearance of $-/+\infty$ values. Returning to the previous DNN example, the same NN can suffer from exploding gradients in case the parameters are large in a way that their products with the derivative of the sigmoid keep them on the higher side until the gradient value explodes and eventually becomes numerically unstable.\\
\textbf{Verification routine.} One can detect unstable gradient issues by examining the first and the last gradients, which correspond respectively to the last, and the first hidden layer (to make the computation cheaper). More specifically, one can check if their $25^{th}$ percentiles (lower quartiles) are not NaN or less than a minimum threshold and their $75^{th}$ percentiles (upper quartiles) are not Inf or higher then a maximum threshold; this could indicate the presence of diminishing or exploding values through backpropagation. Softer tests can be performed by comparing the ratio of absolute average magnitudes of the last gradient by the first gradient to some predefined tunable thresholds(i.e. $min$ and $max$ in the inequation \ref{ineq:ratio_gradients}), indicating the maximum tolerable growth or decline of back-propagated gradients.
\begin{equation}
\label{ineq:ratio_gradients}
min < \log_{10}\left(\frac{mean(abs(last\_gradients))}{mean(abs(first\_gradient))}\right) < max
\end{equation}
\section{TFCheck : A TensorFlow library for Testing ML programs}\label{library}
To assist users in debugging and testing the code implementation of their TF ML programs, we developed a library implementing the issues detection routines described in Section~\ref{heuristics}. We choose to focus on TF programs because of the popularity of TF in the ML community. Nevertheless, the ideas implemented in this library could be adapted for other frameworks. In the following, we describe each module in more details.
\subsection{Setting Up the Testing Session}
\begin{figure}[t]
\centering
\includegraphics[scale=0.4]{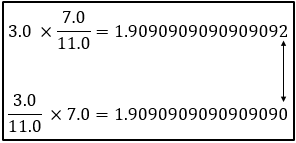}
\caption{Example of result difference caused by operators order change}
\label{diff_parallel_comp}
\vspace{-15pt}
\end{figure}

The testing of a DNN training program can be more complicated than for traditional programs because of its non-deterministic aspect. Indeed, it is difficult to investigate and detect training issues when the program exhibits a novel behavior and returns different output for each execution. However, the testing task can be much easier when we guarantee that the ML program always produces the same outputs given the same inputs.
To avoid stochastic ML program's execution, we analyzed the different aspects of ML implementations that could result in non-deterministic behavior.\\
\textbf{(1) Randomness: }The stochastic nature of the training program is induced by the facts that the model parameters are initialized randomly, mini-batch optimizers select random batches of training data at each iteration until convergence. This results in the final parameters being slightly different between multiple execution times, and recent regularization techniques such as dropout, introduces randomness when it eliminates a number of neurons from training with respect to pre-defined probability.\\
\textbf{(2) Parallelism:} TF relies heavily on parallelism to achieve high performance through multicores CPU and GPU. Multi-threading execution makes it extremely hard to get perfectly reproducible results. To illustrate this point, suppose that two parallel operations (noted $a$ and $b$) are executed simultaneously by two threads, depending on how fast each thread runs, operation $a$ may finish before or after operation $b$. That's not a problem as long as the outputs of these operations are used in a deterministic order, but if they are pushed to the same queue, then the order of the items may change at each single run, and consequently, the downstream operations also change. Although mathematically the result should be the same using real numbers, program may not return the exact same results since the execution environments perform computation using floats with limited precision as illustrated in Figure~\ref{diff_parallel_comp}. During training, these tiny differences will accumulate and generate different outputs in the end. To make ML program deterministic, we first turn off multi-threading execution on CPU and GPU. However, this would be done only for the testing phase since the single thread program execution would be much slower in terms of running time. 
To ensure the reproducibility of the DNN training program under test,
the \emph{setup()} method in our TFCheck library turns off almost all the potential sources of variability by fixing the seeds for all the computational libraries used, such as TF, Numpy and Python built-in random library. TFCheck can also deactivate the parallelism with GPU depending on the usage of the tester, \ie{} whether he prefers more the determinism or the rapidity of execution of the different verification routines in a parallel environment.
\subsection{Monitoring the Training Program Behavior}
TFCheck performs the verifications described in Section~\ref{heuristics} using values of tensors fetched before running training operations and the results obtained after running the training operations. Thus, given the number of verifications that should be performed once in a while, between the \emph{session.run()} calls, we use the monitored session and hooks mechanism to handle all this additional processing injected in the training program. This allows us to keep the code of our library elegant, maintainable, and easy to extend. To develop and use session hooks in our library, we need to perform two steps:
\begin{enumerate}
\item Create one or more Hooks that inherits from the \emph{SessionRunHook} class and implement methods required to perform the check, essentially we implement \emph{before\_run} and \emph{after\_run} that execute pre-and post-processing to each \emph{session.run()} call.
\item Create a \emph{MonitoredSession} and attach to it the implemented session hooks by setting the hooks parameter to a list of session hooks instances.
\end{enumerate}
TFCheck implements all the verification routines described in Section~\ref{heuristics} as well as a \emph{Hooks} module that contains the different session hooks for the different verification routines. Those hook objects access values obtained for activations, parameters, and gradients w.r.t parameters, with the aim of applying the verification routines on them. Since the training program runs consecutive \emph{session.run()} calls, most tensors such as activations and gradients do not survive past a single execution of the graph and the parameters of the model such as weights and biases are updated continuously. Therefore, those hook objects generally store the last values of chosen parameters or previously obtained value to enable the execution of the necessary comparisons. 
\subsection{Logging for any Detected Issue} Once TFCheck finds an issue, it reacts depending on the configuration defined by the tester. The current available configurations are : ``log warning message explaining the detected issue'' or ``stop the testing process and raise an exception''. The \emph{Logging and Exceptions} module contains different exceptions related to the issues tested and meaningful warning messages aimed at helping testers understand the issues of their training programs and helping them identify their root causes.
\section{Evaluation of TFCheck}\label{evaluation}
To assess the effectiveness of TFCheck at detecting errors. We replicate the training of $4$ buggy TensorFlow programs identified by Zhang et al.~\cite{TFBugs} and of $7$ mutants generated and verified by Dwarakanath et al.~\cite{LastMT}. Besides, we create $4$ synthetic training codes that imitate known issues to complement the evaluation of TFCheck. Table \ref{result_table} summarizes the results and following, we describe in detail the issues found and the TFCheck verification routines that detected them. 
\begin{table*}[t]
\label{result_table}
\caption{Overview on the Tested Neural Networks}
\centering
\begin{tabular}{c c c} 
\hline
\textbf{Neural Network} &  \textbf{Issue} & \textbf{Fired Checks}\\
\hline
\textbf{IPS-4} & Inadequate Loss function & Unstable parameters learning(slow), non-decreasing loss \\
\hline
\textbf{IPS-5} & Inefficient Optimization & Unstable parameters learning(high), Exploding gradients  \\
\hline
\textbf{IPS-15} & Poor weight Initialization & Unbreaking Symmetry, Exploding Weights  \\
\hline
\textbf{IPS-17} & High learning rate & Unstable parameters learning(high), Diverging Loss \\
\hline
\textbf{Mutant-29} & Incorrect loss function& Zero Loss  \\
\hline
\textbf{Mutant-30} & Incorrect regularization term& Unstable parameters learning(high), Loss Fluctuations  \\
\hline
\textbf{Mutant-31} & Incorrect regularization term& Unstable parameters learning(high), Loss Fluctuations  \\
\hline
\textbf{Mutant-32} & Incorrect regularization term& Unstable parameters learning(high), Loss Fluctuations   \\
\hline
\textbf{Mutant-43} & Incorrect learning rate schedule& Unstable parameters learning, Vanishing Gradient  \\
\hline
\textbf{Mutant-44} & Incorrect learning rate schedule& Unstable parameters learning, Vanishing Gradient \\
\hline
\textbf{Mutant-45} & Incorrect learning rate schedule& Unstable parameters learning, Vanishing Gradient  \\
\hline
\textbf{Synthetic-1} & Very deep NN with sigmoid activations & Saturated Neurons\\
\hline
\textbf{Synthetic-2} & Random bias initialization with Huge negative values& Dead Neurons\\
\hline
\textbf{Synthetic-3} & Disconnected layers & Untrained parameters  \\
\hline
\textbf{Synthetic-4} & Deactivate a layer (Remove the activation function) & Activation out of range  \\
\hline
\end{tabular}
\vspace{-15pt}
\end{table*}
\subsection{Real-world Training Programs}
In the empirical study on TF Bugs~\cite{TFBugs}, We found programs in which the reported bugs are related to an Incorrect Model Parameter or Structure (IPS). We choose this type of bugs because they often manifest themselves during the training phase of the models, and therefore allow for early detection. These bugs are mainly caused by inappropriate modeling configurations that lead to erroneous behaviors during the training phase. Zhang et al. claim that the major symptom of these bugs is low effectiveness, i.e., low accuracy. However, we believe that TFCheck can generate more fine-grained feedbacks whenever such issue is encountered by ML developers; which will help for its early detection and for understanding its root cause.

To verify our hypothesis, we replicate these buggy TF programs. Then, their executions using a monitored training, incorporated with TFCheck hooks allowed us to detect the existing issues in $4$ programs.\\
In IPS 4, the neural network use inadequate loss function that, first, triggers an \textit{unstable parameters learning} issue; because parameters seem to be changing slowly and the model seems unable to learn patterns. Over time, this caused the \textit{non-decreasing of the loss} issue and its stagnation at a relatively high value.\\
In IPS 5, the predict variable is a continuous variable and the loss function is MSE. However, the sample training data used contained outputs that are relatively big and the use of gradient descent optimizer with a high learning rate (i.e., $0.1$) caused the \textit{unstable parameters learning} issue in the sense that weights changed wildly with relatively high update steps, then, the \textit{exploding of gradients} issue occured.\\
In IPS 15, the weights are poorly initialized in a way that contain a constant value, which is not small enough to not diverging through backpropagation of gradients. Therefore, TFCheck displayed the \textit{Unbreaking Symmetry} issue from the first iteration. Then, it triggerred the \textit{exploding weights} issue when one of the DNN's weights starts containing huge values.\\
In IPS 17, the learning rate was high in a way that caused the \textit{unstable parameters learning} issue, with remarkable bigger update weights ratio and, then, this ended up turning into a \textit{diverging loss} issue.
\textbf{These results suggest that TFCheck can successfully help developers detect training issues; resulting from misconception or poor choices of DNN's configuration spots.}
\subsection{Mutants of Training Programs}
Dwarakanath et al.~\cite{LastMT} created mutants of training programs that represent a TF training program with one bug. They adopted a systematic approach of changing a line of code, at once, in the original source code and repeatedly generated multiple source code files with intentionally induced faults. To perform a large-scale generation, they used the MutPy tool~\cite{MutPy}, from the Python Software Foundation, to generate the mutants. From their mutants analysis and dataset, we extracted non-crashing mutants that contain a subtle bug in the training code in order to see if TFCheck can trigger adequate warnings for each issue.\\
In Mutant-29, a random change of the mathematical operator in the loss function causes its deficiency. TFCheck was able to kill this mutant during the testing process when it triggered the \textit{Zero Loss} issue warning, indicating that the loss reached a zero value, which is a suspicious value. In fact, if we continue the training execution, the mutant will even return negative values. In Mutant-30-31-32, different deformations in the regularization term causes its defectiveness in a way that it becomes a non-regularization term that would guide the optimizer to increase the value of weights in opposite with regularization objective. TFCheck identified this training issue by, first, alerting for the \textit{Unstable parameters learning} issue caused by high weights' updates steps and second, triggering a warning related to the presence of the \textit{high fluctuations on the loss} issue; showing that the optimizer encounters a difficulty finding the minimum under the given circumstances. In Mutant-43-44-45, some intentionally induced faults in the formula that generates the schedule of different learning rates enhance the weirdness of computed values and the deficiency of the corresponding training session. For this mutant, TFCheck was able to detect those incorrect learning rate schedules through the resulting \textit{unstable parameters learning} verification routine. 
Besides, TFCheck alerted also for the \textit{vanishing gradient} issue when the gradients became smaller and smaller, to adapt the update step given those high learning rates.\\
\textbf{These results show that TFCheck can successfully detect training issues caused by coding mistakes.}
\subsection{Synthetic Training Programs}
We complemented the evaluation of TFCheck using 
synthetic code examples. 
For the saturation neurons issue, we constructed a fully connected NN with ten layers including each 100 neurons with Sigmoid as activation function for the MNIST digits classification (i.e., input images $28\times28$ and $10$ predicted classes). This neural network risks the saturation from early training iterations as explained in the Glorot and Bengio~\cite{glorot2010understanding}. 
For the problem of \emph{dead neurons}, we also implement a ReLU-based fully connected NN that contains a biased initializer that introduces huge negative bias in random neurons. This NN is at risk of dead neurons, since their linear computation could give almost zero or less, because of the huge negative bias. Inspired from a buggy TF code snippet found in github\footnote{\url{https://gist.github.com/Thenerdstation/1c333b33e587859e37476109b44491c1\#file-convnet-py}}, we developed a training program that contains disconnected layers from others in the computation graph; so they did not depend on the training operation. TFCheck successfully detected these known mistakes (it reported about the existence of parameters that are permanently untrained by the DNN during the training session). Regarding activation layer issues, we removed the activation function from the layer definition in order to mimic the situation when the outputs of a newly designed activation function are not as expected. TFCheck triggered the \textit{activation are out of their valid range} warning. 
\textbf{These results further reinforce previous findings that TFCheck can successfully detect training issues in ML programs.}
\section{CONCLUSION}\label{conclusion}
In this paper, we 
introduce a list of verification routines that can be used by developers to detect errors in the implementation of ML models. We implemented these verification routines in an automated testing library for ML programs developed using TensorFlow. 
Evaluation results show that using our library, developers can successfully detect training issues in their ML program implementations. In the future, we plan to expand this library to include the verification of 
advanced activation and loss functions through property-based testing~\cite{OurPaper} (to validate their conformity), and numerical-based testing~\cite{OurPaper} (to check the correctness of their gradient automatically inferred by automatic differentiation).
\balance
\bibliography{main}{}
\bibliographystyle{IEEEtran}

\end{document}